\documentclass[letterpaper, 10 pt, conference]{ieeeconf}  %

\IEEEoverridecommandlockouts                              %

\usepackage[T1]{fontenc}
\usepackage{booktabs}
\usepackage{multirow}
\usepackage{graphicx}
\usepackage[table, dvipsnames]{xcolor}
\usepackage{tikz}
\usetikzlibrary{positioning, arrows.meta, calc, backgrounds, fit}
\usepackage{amsmath}
\usepackage{amsfonts}
\usepackage{rotating} 
\usepackage{comment}
\usepackage{subcaption}

\usepackage{array}
\newcolumntype{C}{>{\centering\arraybackslash}c}

\usepackage[table]{xcolor}
\definecolor{intraavg}{RGB}{235,242,250}
\definecolor{interblock}{RGB}{239,248,242}
\definecolor{interavg}{RGB}{225,240,229}
\definecolor{oursgreen}{RGB}{242,248,242}

\title{\LARGE \bf
Visual Place Recognition in Forests with Depth-Aware Distillation}

\author{Walter Nedov$^{1,2}$, Saimunur Rahman$^{1}$, Kavindie Katuwandeniya$^{1}$, \\ David Hall$^{1}$, Kaushik Roy$^{1}$, and  Peyman Moghadam$^{1,3}$%
\thanks{$^{1}$ CSIRO Robotics, Brisbane, Australia}%
\thanks{$^{2}$ University of Queensland, Brisbane, Australia}%
\thanks{$^{3}$ Queensland University of Technology, Brisbane, Australia}%
}

\usepackage{fancyhdr}
\fancypagestyle{withfooter}{
  \fancyhf{} %

  \fancyfoot[C]{\footnotesize Accepted to the IEEE ICRA Workshop on Field Robotics 2026}
}

\begin{document}

\maketitle

\thispagestyle{withfooter}
\pagestyle{withfooter}

\begin{abstract}

Visual place recognition in natural forest environments remains challenging due to repetitive vegetation, weak structural cues, and significant appearance variation across traversals.
To address this limitation, this paper proposes a lightweight depth-aware distillation framework that injects geometric cues into a DINOv2-based place recognition model, while maintaining its pre-trained descriptor space.
Evaluated on the recent WildCross benchmark, the proposed approach yields gains over an appearance-only counterpart, providing robustness to appearance variations. 
These results demonstrate the importance of depth as a strong complementary modality for place recognition in natural environments and identify depth-aware distillation as a promising direction for more robust forest perception.

\end{abstract}


\section{Introduction}
Visual Place Recognition (VPR) in natural environments remains difficult because the visual evidence available for matching is often weak, repetitive, and unstable across revisits. Forests and other unstructured scenes usually contain few clear landmarks, and the same place can look quite different when seen again under different lighting, vegetation growth, weather, or viewpoint. 
This has recently been made apparent through the new WildCross benchmark~\cite{knights2026wildcross}, a dataset in Australian forests containing aligned RGB and semi-dense depth images with accurate pose information in large-scale natural scenes.

WildCross showed that, in natural environments, visual place recognition (VPR) performs substantially worse than LiDAR place recognition (LPR), where LPR demonstrated an approximately 20-40\% performance advantage~\cite{knights2023wild}, indicating that visual features alone are insufficiently robust for reliable place recognition in such settings.

Current VPR systems have shown their best performance in urban environments, with steady improvements increasingly driven by vision foundation models. CosPlace~\cite{berton2022rethinking} advanced RGB-based retrieval through stronger metric learning and improved scalability, while DINOv2~\cite{oquab2023dinov2} provided robust and transferable visual features across tasks and image domains. More recently, SALAD~\cite{izquierdo2024optimal} and Pair-VPR~\cite{hausler2025pair} have emerged as state-of-the-art methods that leverage pre-trained transformer representations to produce more discriminative place descriptors and stronger matching performance.

\begin{figure}
    \centering
    \includegraphics[width=1\linewidth]{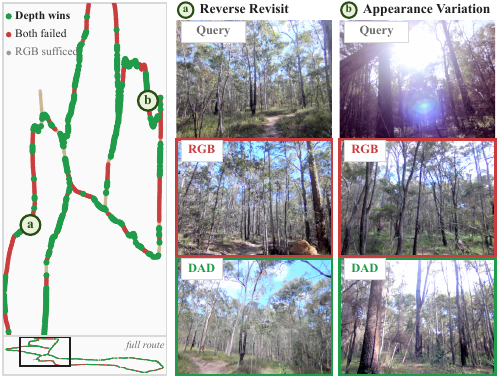}
    \caption{Depth provides complementary cues for natural scene place recognition. Left: query locations where the depth-augmented model outperforms the appearance-only baseline on a challenging reverse traversal. Right: two representative examples. In both cases, RGB only retrieval is distracted by visually similar appearance, whereas the depth-augmented model retrieves a more geometrically consistent match. This behavior motivates our DAD framework.}
    \label{fig:hero-figure}
\end{figure}

However, VFMs can still suffer in VPR systems due to the visual characteristics of natural scenes. 
Repeated vegetation, weak structural cues, and viewpoint variations can make different places look similar and the same place look different. Fig. \ref{fig:hero-figure} illustrates this failure mode. In both examples, the RGB only model retrieves scenes that look reasonable at first glance but are not the best match for the true place. 
Errors such as this motivate our research into other modalities to complement vision for place recognition.

Depth images are a promising avenue for this research, providing geometric information about the shape and layout of the scene in a format that matches that of colour images. 
When the appearance is ambiguous, geometry can provide a useful extra cue. This is also visible in Fig. \ref{fig:hero-figure}, where the depth-augmented model retrieves a better match than the RGB-only model. 
In our work, we examine how depth data can complement strong pre-trained VPR models to improve image descriptors without damaging the model structure that makes retrieval effective.

\begin{figure*}[t]
    \centering
    \includegraphics[width=1\linewidth]{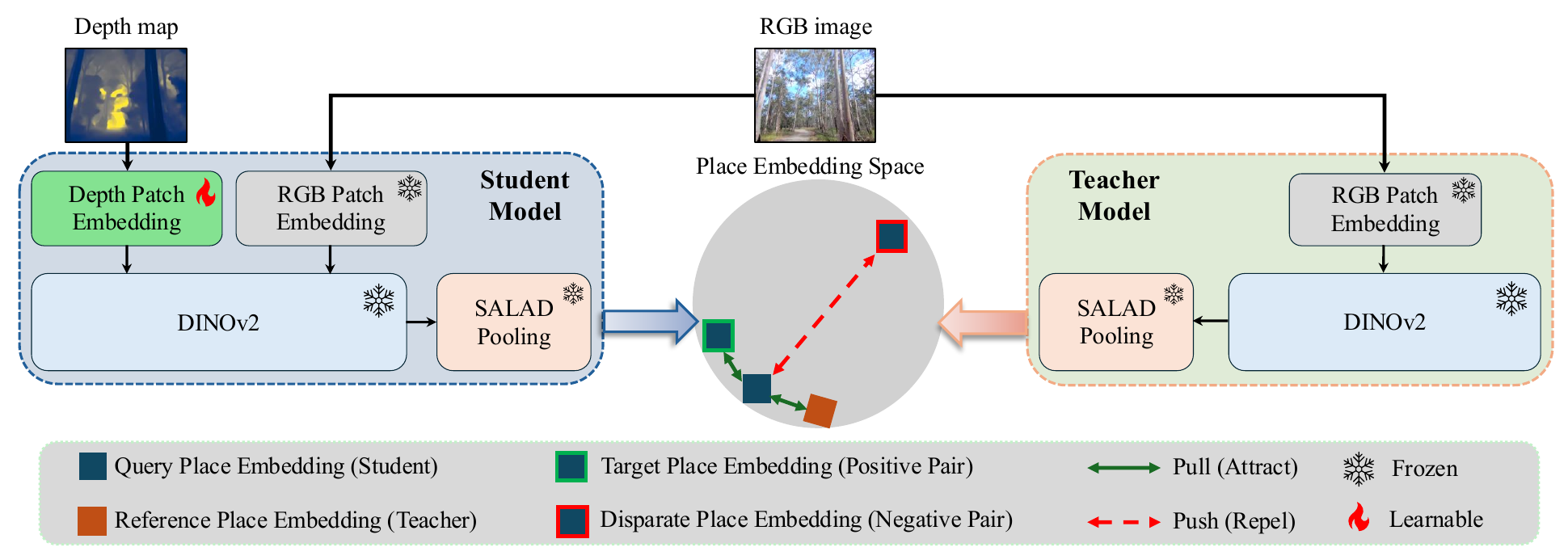}
    \caption{Proposed depth-aware distillation (DAD) framework. The frozen appearance-only teacher maps RGB input to a global descriptor $\mathbf{z}^t$. The student augments the same pretrained visual pathway with a learnable depth embedding, producing a depth-aware descriptor $\mathbf{z}^s$. Training jointly optimizes a triplet retrieval loss on $\mathbf{z}^s$ with a descriptor-level alignment loss between $\mathbf{z}^s$ and $\mathbf{z}^t$. The triplet loss pulls embeddings of similar places closer while pushing those of dissimilar places apart, whereas the alignment loss simultaneously encourages the student embedding to remain consistent with its corresponding teacher embedding. %
    Best viewed in color.}
    \label{fig:main-framework}
\end{figure*}

To that end, we present the novel \underline{d}epth-\underline{a}ware \underline{d}istillation (DAD) framework. 
Rather than relearning descriptors with RGB and depth jointly optimized from scratch, DAD adapts the visual features through a learnable depth patch embedding.
DAD is organized as a teacher-student framework. The teacher is an appearance-only branch that remains frozen and preserves the descriptor structure learned by the pretrained VPR model. The student augments that same representation with depth and learns how to use geometry to improve retrieval. Together, they serve a simple purpose. Depth should refine a strong appearance based descriptor without disrupting the structure that already makes it effective. This choice is also well suited to the modality itself. Unlike LiDAR point clouds, depth remains image-aligned and preserves scene structure in the same spatial frame as RGB, which makes it a natural source of complementary geometric information for refining an appearance based descriptor.

Training is guided by two complementary objectives. A ranking loss (i.e., triplet loss) makes the student descriptor more discriminative for place recognition by pulling matching places closer and pushing non-matching places apart. An alignment loss then distills the teacher into the student by keeping the student descriptor close to the one produced by the frozen appearance-only branch. These two objectives play different roles. The ranking term encourages the student to improve retrieval, while the alignment term prevents the depth-augmented representation from drifting too far away from the pretrained visual prior. By doing so, DAD does not treat depth as a fully separate modality with its own independent descriptor space. Instead, it learns a depth patch embedder whose role is to inject geometry into a strong pre-trained visual descriptor while preserving the appearance based structure that already supports good matching.

The contribution of this work is twofold. First, we introduce DAD, a simple framework that refines a strong pretrained appearance-based VPR descriptor with depth while preserving the retrieval structure already encoded by the underlying visual model. In our implementation, this descriptor is built on DINOv2. Second, we demonstrate that selectively fine-tuning the depth pathway alone, without modifying the pre-trained appearance backbone, yields more discriminative descriptors and retrieval performance competitive with a fine-tuned RGB baseline under zero-shot transfer evaluation, demonstrating that depth can provide useful complementary information for natural-scene VPR.

\section{Proposed method}

Our approach is based on the idea that depth should complement, rather than overwrite, a strong pre-trained appearance-based VPR descriptor. We therefore formulate the DAD framework with a frozen appearance-only teacher and a depth-augmented student. The teacher preserves the descriptor structure learned by the pre-trained VPR model, while the student learns to inject geometric cues for retrieval without drifting far from that descriptor space.

\begin{table*}[t]
\centering
\caption{Retrieval results on WildCross sequences. Left: intra-sequence results for each traversal, followed by the intra-sequence average. Right: inter-sequence results on Venman and Karawatha, followed by the inter-sequence average. }
\label{tab:zeroshot_combined}
\resizebox{\textwidth}{!}{%
\begin{tabular}{
l
*{18}{C}
@{\hspace{2.0em}}
*{6}{C}
}
\toprule
& \multicolumn{18}{c}{\textbf{Intra-Sequence}}
& \multicolumn{6}{c}{\textbf{Inter-Sequence}} \\
\cmidrule(r{1.0em}){2-19}
\cmidrule(l{1.0em}){20-25}
\multirow{2}{*}{\textbf{Method}}
& \multicolumn{2}{c}{\textbf{V-01}}
& \multicolumn{2}{c}{\textbf{V-02}}
& \multicolumn{2}{c}{\textbf{V-03}}
& \multicolumn{2}{c}{\textbf{V-04}}
& \multicolumn{2}{c}{\textbf{K-01}}
& \multicolumn{2}{c}{\textbf{K-02}}
& \multicolumn{2}{c}{\textbf{K-03}}
& \multicolumn{2}{c}{\textbf{K-04}}
& \multicolumn{2}{c}{\textbf{Average}}
& \multicolumn{2}{c}{\textbf{Venman}}
& \multicolumn{2}{c}{\textbf{Karawatha}}
& \multicolumn{2}{c}{\textbf{Average}} \\
\cmidrule(lr){2-3}   \cmidrule(lr){4-5}   \cmidrule(lr){6-7}   \cmidrule(lr){8-9}
\cmidrule(lr){10-11} \cmidrule(lr){12-13} \cmidrule(lr){14-15} \cmidrule(lr){16-17}
\cmidrule(r{1.0em}){18-19}
\cmidrule(l{1.0em}){20-21}
\cmidrule(lr){22-23} \cmidrule(lr){24-25}
& R1 & R5  & R1 & R5  & R1 & R5  & R1 & R5
& R1 & R5  & R1 & R5  & R1 & R5  & R1 & R5
& R1 & R5  & R1 & R5  & R1 & R5  & R1 & R5 \\
\midrule
Baseline
& 50.64 & 57.01
& 44.60 & 49.06
& 13.05 & 19.58
& 43.72 & 48.53
& 78.48 & 83.50
& 54.71 & 68.16
& 26.67 & 32.78
& 54.93 & 59.31
& 45.85 & 52.24
& 57.49 & 64.49
& 41.27 & 50.14
& 49.38 & 57.32 \\
Baseline fine-tuned
& \textbf{68.10} & \textbf{72.40}
& \textbf{72.20} & \textbf{76.80}
& 23.60 & 31.02
& 59.80 & 64.40
& 86.30 & \textbf{88.50}
& 84.40 & 90.00
& 30.00 & \textbf{38.20}
& \textbf{59.80} & \textbf{62.00}
& 60.50 & 65.40
& 66.60 & 71.01
& 49.68 & 55.81
& 58.14 & 63.41 \\
\rowcolor{oursgreen}
DAD (Ours)
& 64.00 & 68.66
& 70.12 & 75.67
& \textbf{28.05} & \textbf{37.78}
& \textbf{60.65} & \textbf{68.39}
& \textbf{86.63} & 87.89
& \textbf{88.52} & \textbf{93.38}
& \textbf{30.27} & 35.86
& 58.35 & 59.85
& \textbf{60.82} & \textbf{65.94}
& \textbf{68.35} & \textbf{73.70}
& \textbf{57.04} & \textbf{65.67}
& \textbf{62.69} & \textbf{69.68} \\
\bottomrule
\end{tabular}%
}
\end{table*}

Both teacher and student are initialized from a pre-trained SALAD model \cite{izquierdo2024optimal} with a DINOv2 encoder. In this initialization, DINOv2 \cite{oquab2023dinov2} provides the visual feature extractor, and SALAD provides the global descriptor aggregation module. Fig.~\ref{fig:main-framework} gives an overview of the framework. Let $I$ denote an RGB image and $D$ its corresponding depth map, obtained using the pre-trained Depth Anything V2 model \cite{yang2024depth}. We denote by $F_v(\cdot)$ the visual encoder and by $G(\cdot)$ the descriptor aggregation function inherited from the pre-trained model. The teacher maps the RGB input to a global descriptor,
\begin{equation}
\mathbf{z}^{t} = G(F_v(I)),
\end{equation}
which serves as the reference appearance-based representation. The teacher branch remains frozen throughout training.

The student shares the same pre-trained visual encoder and descriptor aggregation module, but introduces a learnable depth embedding pathway $F_d(\cdot)$ to encode geometric cues from the depth map. The depth path mirrors the RGB patch-embedding architecture, except that it operates on a single-channel input. The student descriptor is defined as
\begin{equation}
\mathbf{z}^{s} = G\!\left(F_v(I) + \alpha F_d(D)\right),
\label{eq:student_descriptor}
\end{equation}
where $\alpha$ is a learnable scalar that modulates the contribution of depth. Under this formulation, depth acts as a geometry-conditioned correction to the pre-trained appearance representation, rather than as a separate modality-specific encoder. During training, only the depth embedder $F_d(\cdot)$ and the scaling parameter $\alpha$ are optimized, while the pre-trained visual components remain frozen. We initialize $\alpha$ to $0.01$.

Training is guided by two complementary objectives that jointly enforce discriminability and geometric consistency in the descriptor space. The first is a triplet loss that makes the student descriptor discriminative for place recognition. Given an anchor-positive-negative tuple, with corresponding student descriptors $\mathbf{z}_a^{s}$, $\mathbf{z}_p^{s}$, and $\mathbf{z}_n^{s}$, the retrieval loss is
\begin{equation}
\mathcal{L}_{\mathrm{triplet}}
=
\max\!\left(
0,\,
m + d(\mathbf{z}_a^{s},\mathbf{z}_p^{s}) - d(\mathbf{z}_a^{s},\mathbf{z}_n^{s})
\right),
\label{eq:triplet_loss}
\end{equation}
where $d(\cdot,\cdot)$ denotes descriptor distance and $m>0$ is a margin. This term encourages descriptors from the same place to lie closer than those from different places.

The second objective is a descriptor-level alignment loss that distills the teacher representation into the student. Specifically, we define $\mathcal{L}_{\mathrm{align}} = \ell(\mathbf{z}^{s},\mathbf{z}^{t})$, where $\ell(\cdot,\cdot)$ measures the discrepancy between student and teacher descriptors. Using cosine distance, this becomes
\begin{equation}
\mathcal{L}_{\mathrm{align}}
=
1-\frac{\langle \mathbf{z}^{s},\mathbf{z}^{t}\rangle}
{\|\mathbf{z}^{s}\|_2 \, \|\mathbf{z}^{t}\|_2}.
\label{eq:align_loss}
\end{equation}

By reducing angular discrepancy, this term enforces directional consistency between $\mathbf{z}^s$ and $\mathbf{z}^t$, aligning the student with the teacher’s embedding.
The overall training objective is
\begin{equation}
\mathcal{L} = \mathcal{L}_{\mathrm{triplet}} + \lambda\mathcal{L}_{\mathrm{align}},
\label{eq:total_loss}
\end{equation}
where $\lambda$ balances retrieval supervision and teacher alignment. We fix $\lambda$ at 0.05. At inference time, retrieval is performed using the student descriptor $\mathbf{z}^{s}$.

Geometrically, the two loss terms play distinct yet complementary roles in descriptor space. The triplet loss shapes the local retrieval geometry by pulling descriptors of matching places closer and pushing non-matching places apart, thereby improving discriminability. The alignment loss, in contrast, anchors the student to the teacher-defined descriptor space, preserving compatibility with the pretrained appearance-based prior while depth is introduced.

\section{Results and Discussion}

\begin{figure}[hbt!]
    \centering
    \includegraphics[width=\columnwidth, height=0.3\textheight, keepaspectratio]
{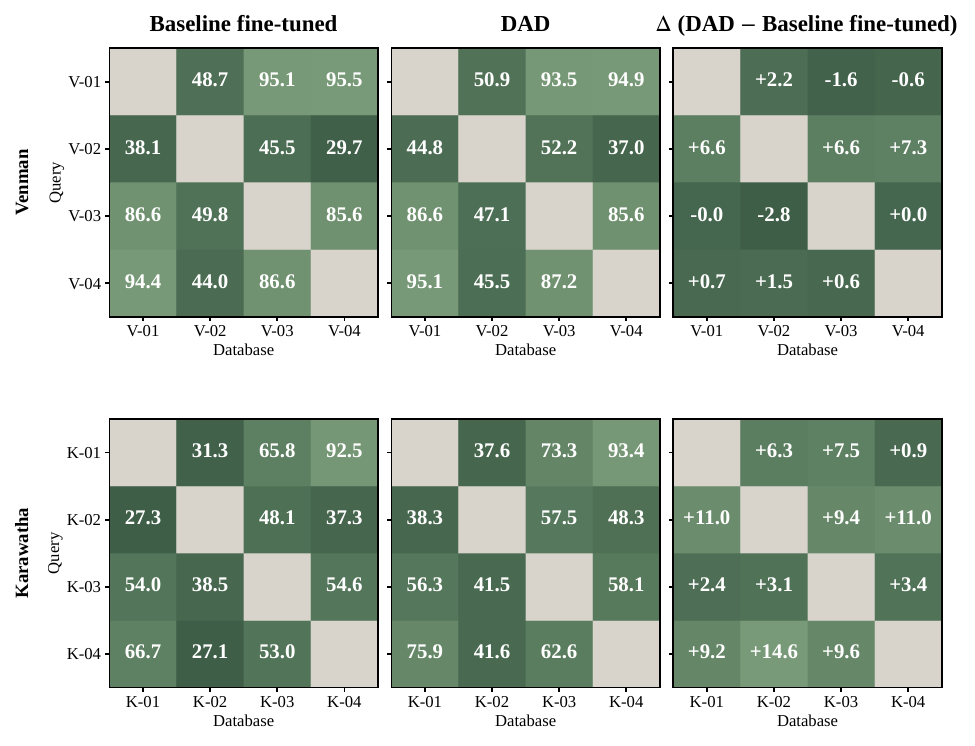}
    \caption{Inter-sequence R@1 heatmaps for zero-shot evaluation on WildCross. Rows and columns correspond to query and database sequences, respectively. Diagonal cells are empty (intra-sequence pairs excluded). Baseline fine-tuned (left), DAD (centre), and their difference (right).}
    \label{fig:heatmap_zeroshot}
\end{figure}

\paragraph*{Evaluation}
We evaluate DAD on WildCross, which is well-suited to this study because it targets place recognition in natural environments and provides paired RGB and depth observations for the same scenes. In our experiments, we use the depth images released with WildCross, which are generated using the Depth Anything V2 model. As the underlying RGB descriptor pipeline, we use a pretrained SALAD checkpoint, with DINOv2 as the visual backbone and SALAD as the global descriptor pooling module. During training, this pretrained RGB pathway remains fixed, and only the depth patch embedder and the depth scaling parameter are optimized. We compare against two baselines: the frozen pre-trained model (Baseline) and a variant with the RGB patch embedder fine-tuned (Baseline fine-tuned), with all other components held fixed. Following the WildCross protocol, all results are reported in the zero-shot setting \cite{knights2026wildcross}. This protocol is particularly relevant here because it tests generalisation to unseen sequences without sequence-specific fine-tuning of the appearance pathway.

The main quantitative results are summarized in Table \ref{tab:zeroshot_combined}. DAD improves over the appearance-only SALAD baseline in both intra-sequence and inter-sequence evaluation. For intra-sequence retrieval, the average Recall@1 improves from 45.85 to 60.82, and Recall@5 improves from 52.24 to 65.94. For inter-sequence retrieval, the average Recall@1 improves from 49.38 to 62.69, and Recall@5 improves from 57.32 to 69.68. The gains are also visible in both Venman and Karawatha, which shows that the benefit is not limited to one environment split. These results support the main claim of the paper. A strong pre-trained appearance-based descriptor can be improved meaningfully by depth, even when the depth cue is obtained from predicted monocular depth rather than native 3D sensing.

Fig. \ref{fig:heatmap_zeroshot} gives a more detailed view through the inter-sequence heatmaps for RGB, DAD, and their difference. Two patterns are clear. First, the improvement is widespread rather than isolated. Most inter-sequence pairs benefit from depth. Second, the gains are not uniform. Some traversal pairs improve more than others, which suggests that depth is especially useful when retrieval is difficult because appearance alone is ambiguous. This behavior is consistent with the design of DAD. The method does not replace the original descriptor space with a new multimodal embedding learned from scratch. Instead, it refines the pretrained appearance-based descriptor with geometric information while remaining aligned with the teacher.

\begin{figure}
    \centering
    \includegraphics[width=\columnwidth, height=0.26\textheight]{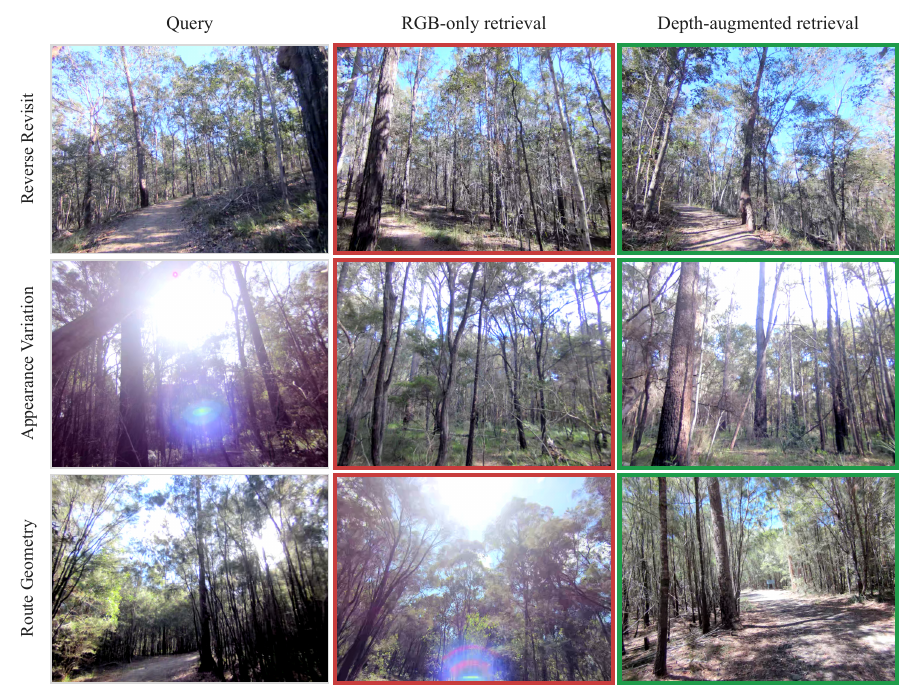}
    \caption{Example cases where DAD improves retrieval. From left to right, each row shows the query, the incorrect RGB-only retrieval, and the corrected depth-augmented retrieval produced by DAD. Red and green borders indicate false retrievals and correct retrievals, respectively. }
    \label{fig:where-depth-helps}
\end{figure}

The qualitative examples help explain why depth improves retrieval. Fig.~\ref{fig:hero-figure} illustrates the central observation of the paper, while Fig.~\ref{fig:where-depth-helps} shows additional representative failure cases. In the same cases where the RGB-only model retrieves visually plausible but incorrect matches, DAD retrieves a better match by using depth as a complementary cue. Across these examples, three patterns appear repeatedly. First, depth helps in cases of scale ambiguity, where nearby trunks, open path regions, or surrounding vegetation look similar in RGB but differ in spatial extent. Second, depth helps when a large foreground object is close to the camera, which can distract RGB-only retrieval toward visually similar but incorrect matches. Third, depth helps when path geometry is distinctive, especially where the path curves or recedes in a way that is only weakly captured by texture or color. In these cases, depth provides structural evidence that remains useful when appearance alone is confusing.

\begin{figure}
    \centering
    \includegraphics[width=1\linewidth]{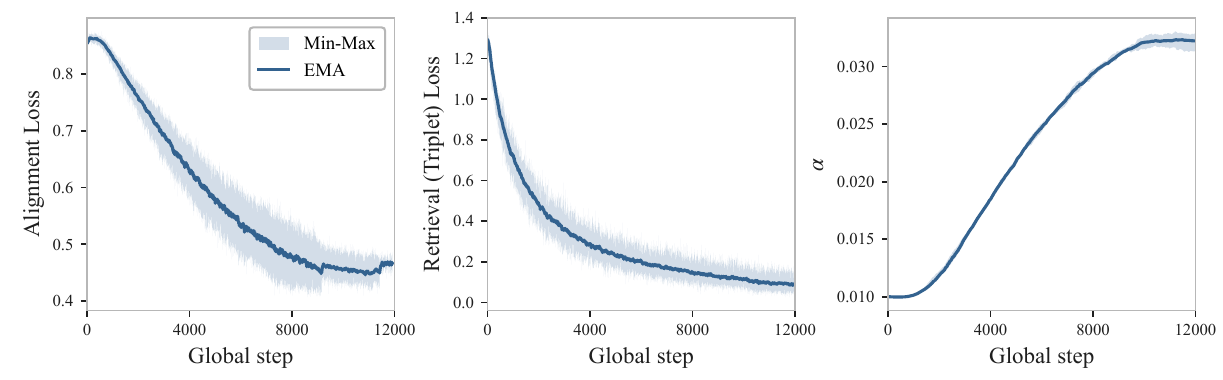}
    \caption{Training curves of DAD. The alignment and retrieval losses decrease over training, while the learned depth scale $\alpha$ stabilizes after moving away from its initialization.}
    \label{fig:loss}
\end{figure}

Fig. \ref{fig:loss} shows stable optimization of DAD. Both the alignment loss and retrieval loss decrease over training, while the learned depth scale $\alpha$ moves away from its initialization and then stabilizes. This indicates that the student becomes more discriminative for retrieval while using depth in a controlled way. Together, these trends support the role of the two objectives in DAD: the retrieval loss improves place discrimination, while the alignment loss preserves compatibility with the pre-trained appearance based descriptor space.

\section{Conclusion and Future Work}

We presented DAD, a simple depth-aware distillation framework for improving VFM based VPR in natural environments. DAD refines a strong pre-trained appearance-based descriptor with depth while preserving the retrieval structure already encoded by the visual model. On WildCross, DAD produced clear gains over the appearance-only baseline in both intra-sequence and inter-sequence zero-shot retrieval. The qualitative results showed that depth is especially helpful when appearance is ambiguous, such as in cases of reverse revisit and appearance variation. Future work will evaluate DAD with LiDAR-derived geometry and test whether the approach extends to other backbones and unstructured benchmarks.

\bibliographystyle{ieeetr}
\bibliography{references.bib}

\end{document}